\documentclass[times,twocolumn,final]{elsarticle}

\usepackage{medima}
\usepackage{framed,multirow}

\usepackage{amssymb}
\usepackage{latexsym}

\usepackage{url}
\usepackage{xcolor}
\usepackage{amsmath}
\usepackage{algorithmic}
\usepackage{algorithm}
\usepackage{booktabs} 

\usepackage{hyperref}

\definecolor{newcolor}{rgb}{.8,.349,.1}

\journal{Medical Image Analysis}

\begin{document}

\verso{Jiawen Li \textit{et~al.}}

\begin{frontmatter}

\title{Diagnostic Text-guided Representation Learning in Hierarchical Classification for Pathological Whole Slide Image}%

\author[1,3]{Jiawen Li}              
\author[1,3]{Qiehe Sun}               
\author[1]{Renao Yan}                      
\author[1,3]{Yizhi Wang}
\author[1]{Yuqiu Fu}
\author[2]{Yani Wei}      
\author[1]{Tian Guan}
\author[2]{Huijuan Shi\corref{cor1}}
\author[1,3]{Yonghong He\corref{cor1}}
\author[2]{Anjia Han\corref{cor1}}
\cortext[cor1]{Corresponding author}

\ead{shihj@mail.sysu.edu.cn, heyh@sz.tsinghua.edu.cn, hananjia@mail.sysu.edu.cn}

\address[1]{Shenzhen International Graduate School, Tsinghua University, Shenzhen 518055, China}
\address[2]{Department of Pathology, The First Affiliated Hospital of Sun Yat-sen University, Guangzhou 510080, China}
\address[3]{Medical Optical Technology R\&D Center, Research Institute of Tsinghua, Pearl River Delta, Guangzhou 510700, China}




\begin{abstract}
With the development of digital imaging in medical microscopy, artificial intelligent-based analysis of pathological whole slide images (WSIs) provides a powerful tool for cancer diagnosis. Limited by the expensive cost of pixel-level annotation, current research primarily focuses on representation learning with slide-level labels, showing success in various downstream tasks. However, given the diversity of lesion types and the complex relationships between each other, these techniques still deserve further exploration in addressing advanced pathology tasks. To this end, we introduce the concept of hierarchical pathological image classification and propose a representation learning called PathTree. PathTree considers the multi-classification of diseases as a binary tree structure. Each category is represented as a professional pathological text description, which messages information with a tree-like encoder. The interactive text features are then used to guide the aggregation of hierarchical multiple representations. PathTree uses slide-text similarity to obtain probability scores and introduces two extra tree-specific losses to further constrain the association between texts and slides. Through extensive experiments on three challenging hierarchical classification datasets: in-house cryosectioned lung tissue lesion identification, public prostate cancer grade assessment, and public breast cancer subtyping, our proposed PathTree is consistently competitive compared to the state-of-the-art methods and provides a new perspective on the deep learning-assisted solution for more complex WSI classification.
\end{abstract}


\end{frontmatter}


\section{Introduction}
Modern computational imaging technology has promoted the transformation from the traditional microscope-based to the digital imaging analysis paradigm in pathology (\citealt{niazi2019digital,song2023artificial}). Through advanced optical imaging systems, pathological tissue sections are rapidly converted into gigapixel-level WSIs, preserving the entire tissue structure and lifting the limitations imposed by the fixed field of view during the examination. The emergence of WSIs improves the work efficiency of pathologists and promotes the application of computer vision technology in diagnosing pathology (\citealt{li2022comprehensive,perez2024guide}).

With the efficient utilization of clinical data and computational resources, extensive research has been conducted based on deep learning for WSI analysis (\citealt{rodriguez2022artificial,song2023artificial}). Due to the considerable size characteristic, existing high-performance models designed for analyzing much smaller natural images cannot be directly transferred to high-resolution pathological images. Previous studies are mainly based on patch-level learning (\citealt{wang2016deep,coudray2018classification,yang2021deep,li2023deeptree}), which follows the pipeline as such. Starting by segmenting the WSI foreground into several patches with appropriate sizes, each patch is assigned a manual label to train an encoder and predictor. However, limited by the challenge of acquiring large-scale pixel-level manual annotations with domain-specific expertise, research has shifted from original patch-level supervision toward slide-level representation learning (\citealt{campanella2019clinical,lu2021data,song2024morphological,jaume2024transcriptomics}). These methods use a pre-trained feature extractor to obtain local embeddings and then integrate them through a learnable aggregation module to serve as the WSI feature. Such a global representation can accomplish the target optimization through different slide-level supervision signals.

Current WSI representation learning has demonstrated success in common slide-level tasks. For example, algorithms based on multiple instance learning (\citealt{li2021dual,zhang2022dtfd,xiang2022exploring}), graph representation (\citealt{guan2022node,chen2021whole,li2024dynamic}), transformer (\citealt{chen2022scaling,shao2021transmil,xiong2023diagnose}) and their variants (\citealt{yang2024mambamil,chu2024retmil}) show excellent performance in tasks such as cancer detection for sentinel lymph nodes (\citealt{bejnordi2017diagnostic,litjens20181399,bandi2018detection}) and non-small cell lung cancer subtyping. However, there are three notable points regarding these tasks. First, the number of categories involved in these classification tasks is typically small, usually limited to two or three categories. Second, there is a marked distinction between categories, such as non-cancerous versus cancerous tissues or well-differentiated adenocarcinoma versus squamous cell carcinoma. Third, the task difficulty is relatively low. These datasets represent fundamental issues in pathology, often encountered in the routine and repetitive work of pathologists. 

For challenging pathology tasks such as judging the invasiveness of cancer or detailed subtyping related to prognosis (\citealt{echle2021deep}), the high complexity of lesion morphology and the variability in subjective judgments among different pathologists make it struggle to extract meaningful representations from such highly heterogeneous data. To illustrate this heterogeneity, pathologists use hierarchical structures to distinguish categories step by step (\citealt{kundra2021oncotree}), but existing models do not take this into account. In addition, the higher the difficulty of the pathology task is, the more difficult it will be for a simple text word or phrase to describe a category accurately. Therefore, richer text information is needed to explain the observed diagnostic phenomenon (\citealt{zhang2019text}), thereby guiding and optimizing the WSI representation with adequate information. Although a few studies explore the impact of pathological text on regions of interest or WSI diagnostic analysis (\citealt{li2024generalizable,zhang2023text,shi2024vila,10155265}), they have yet to conduct in-depth discussions on the complex pathological context.

To overcome these challenges, inspired by the authoritative diagnostic strategies of pathologists, we transform the challenging pathological multi-classification issues into hierarchical classification tasks with binary tree structures and design a hierarchical WSI representation learning method called PathTree. Specifically, we describe each category (including coarse-grained and fine-grained nodes in the tree) with professional pathological terminology. Then, these texts are encoded into high-dimensional embeddings and passed through a tree-like graph to exchange messages. For image modality, PathTree aggregates the patch embeddings into multiple slide-level embeddings. These embeddings are passed into the root node according to the tree path, guided by the text embeddings. Unlike the conventional linear classifier, PathTree obtains the prediction scores based on slide-text similarity and introduces two extra tree-like aware losses to capture the semantic matching relationship between hierarchical text embeddings. To validate the superiority of our proposed PathTree, we conduct extensive experiments on three large-scale hierarchical datasets, including internal cryosectioned lung tissue identification, public prostate cancer grading, and public breast cancer subtyping. The results demonstrate the consistently outstanding performance of PathTree. Our proposed hierarchical classification concept and method provide more objective and precise assistance in challenging pathological tasks. 

Our contributions can be summarized as follows.

\begin{itemize}
	\item[1)] We describe the challenging WSI multi-classification as the hierarchical tasks with binary tree structure and propose PathTree to represent and analyze WSIs.
	\item[2)] We utilize the textual modality to guide the formation of slide features and measure the potential semantic relationship between hierarchical texts.
    \item[3)] We design and conduct experiments on three hierarchical pathological WSI classification datasets. The results show the effectiveness of our proposed method. Our code will be available later.

\end{itemize}

\section{Related Work}
\subsection{Representation learning of WSIs}

The multi-level resolution and high heterogeneity make it difficult for deep neural networks to extract WSI representations. Some works are denoted to designing directly trainable vision models to obtain thumbnail representations. For example, \cite{pinckaers2020streaming} combines forward and backward propagation with gradient checkpointing, allowing WSIs of any size into convolutional neural networks (CNNs). \cite{chen2021annotation} incorporates the unified memory mechanism and GPU memory optimization techniques to facilitate the entire WSI into CNNs. \cite{xiang2022dsnet} utilizes multi-scale thumbnails and neural-encoded embeddings to achieve dual-stream information transmission of features. \cite{wang2023image} designs LongViT to capture both short-range and long-range dependencies and generate representations of high-resolution thumbnails.

Beyond end-to-end methods, more researches focus on obtaining global representations from patch-level embeddings, and multiple instance learning (MIL) is one such widely discussed approach (\citealt{laleh2022benchmarking}). It treats cropped patches as instances, the WSI as a bag, and generates bag-level features by aggregating the instance-level scores or embeddings. ABMIL proposed by \cite{ilse2018attention} is one of the most common MIL approaches, where each instance generates an attention score based on its feature information, and the bag representation is obtained by weighting the features of all instances. Building on this, variants of ABMIL, such as CLAM (\citealt{lu2021data}), which utilizes clustering constraints, DSMIL (\citealt{li2021dual}), which employs the multi-scale pyramid for feature fusion, TransMIL (\cite{shao2021transmil}), which uses the correlated instance update strategy and so on (\citealt{zhang2022dtfd,qiehe2024nciemil,xiang2022exploring}), have recently been extensively explored and seen numerous successful applications (\citealt{lu2021ai,fremond2023interpretable,wagner2023transformer}).

MIL-based approaches predominantly focus on the instances themselves but struggle to capture the interactions between these instances. Conversely, graph representation methods treat the WSI as a graph and consider cells, tissues, or patches as nodes to convey local context between entities. For example, \cite{chen2021whole} introduces Patch-GCN and utilizes it to model the significant morphological feature interactions between cell identities and tissue types to generate global WSI representations. \cite{li2023high,di2022generating} obtains a high-order global representation of WSIs via multilateral correlation and a hypergraph convolution network. \cite{chan2023histopathology} designs a heterogeneous graph for WSI analysis by utilizing the relations between different types of nuclei. \cite{li2024dynamic} models the patch-level inter-entity correlations in WSI as directed knowledge graphs for cancer classification and staging.

Finally, WSIs can be modeled and updated as sequence representations with Transformer (\citealt{vaswani2017attention}). \cite{huang2021integration} combines self-supervised learning with Transformers to achieve slide-level survival prediction. \cite{zheng2022graph} brings the graph-transformer into the computational pathology and successfully classifies WSIs of lung tissues. \cite{chen2022scaling} designs a hierarchical image pyramid transformer to model representations from cellular spaces to tissue microenvironment. Furthermore, \cite{chen2021multimodal} also joint-learn WSIs and genomic modalities to design a co-attention transformer that represents survival status.

Unlike the above methods, PathTree directly models the interrelationships between pathological morphologies and utilizes diagnostic text to guide and optimize the global representation.

\subsection{Hierarchical classification of pathological WSIs}

\begin{figure}[tbp]
\centering
\includegraphics[width=1\linewidth]{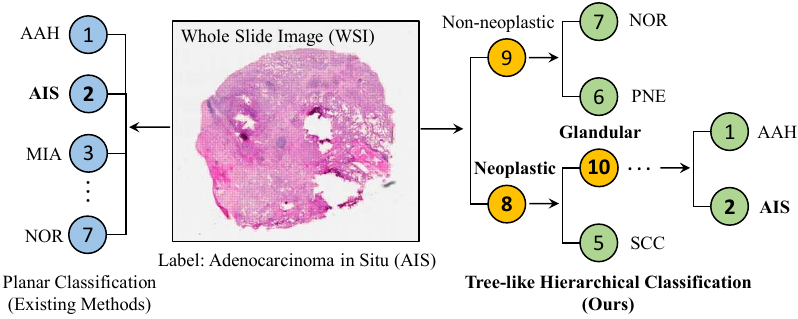}
\caption{The existing WSI multi-classification method is described as a planar classification problem, which treats each category independently and equally. However, the interrelations among various categories exhibit considerable complexity in real-world pathological contexts. Pathologists typically analyze from coarse-grained to fine-grained levels, following a hierarchical, tree-like relationship among classes. This approach enhances the efficiency and precision of diagnostics and helps pathologists systematically understand and interpret complex pathological information, making more accurate medical decisions.}  
\label{flat_vs_hier}
\end{figure}

Previous WSI analysis studies have mainly been conducted on foundational pathology classification tasks. For example, the Camelyon16 dataset for detection of breast cancer lymph node metastasis (\citealt{bejnordi2017diagnostic,bandi2018detection,litjens20181399}), the TCGA-NSCLC for non-small cell lung cancer, and the TCGA-RCC dataset for renal cell carcinoma (\citealt{linehan2019cancer}). However, these binary or ternary disease-state prediction tasks hardly reflect the broader detailed lesions observed in real-world pathological practice (\citealt{chen2024towards}). Inspired by the decision-making processes of pathologists during clinical practice and the OncoTree cancer classification system (\citealt{kundra2021oncotree}), we consider challenging pathological diagnostic issues and model them as hierarchical classification tasks. Fig.\ref{flat_vs_hier} displays the formal differences between planar and hierarchical tree-like analysis structures. Compared to planar methods, tree structures can describe categories from the coarse-grained macro to the fine-grained micro level, for example, from the primary type (such as glandular or squamous) to the degree of cell infiltration (such as non-infiltrated or thoroughly infiltrated). Our previous work (\citealt{li2023deeptree}) uses tree-like deep learning architectures to obtain the representation of pathological regions of interest. In this paper, we focus on slide-level learning since it is closer to clinical applications. In summary, hierarchical classification ensures the comprehensiveness and precision of diagnoses. It aids in detecting early signs or subtle differences in diseases and is crucial for formulating prognostic plans.

\section{Method}

\begin{figure*}[tbp]
\centering
\includegraphics[width=0.89\linewidth]{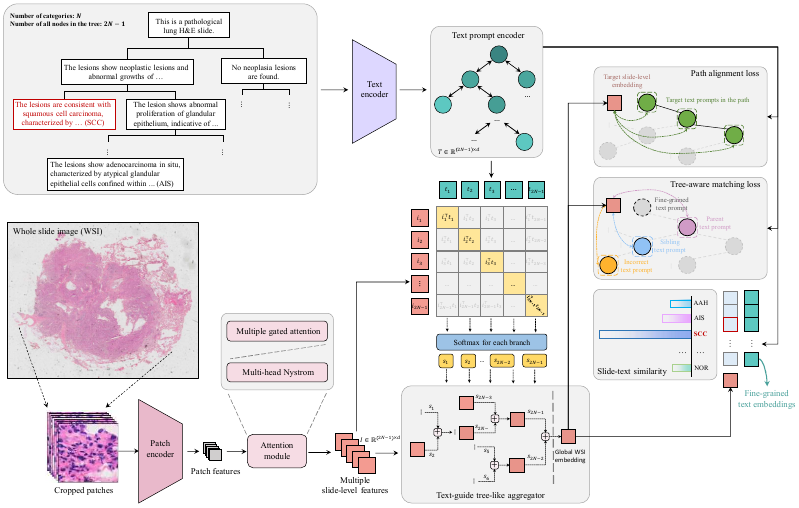}
\caption{Overview of PathTree. The main idea is to convert challenging pathological multi-classification into hierarchical tree structures for analysis. PathTree uses the WSI to generate multiple slide-level embeddings from patch-level features, allowing them to contrast text semantics. After aggregation based on the tree path, the relationship between tree-like text semantics is measured by two structure-specific losses, and prediction scores are obtained by calculating the cosine similarity between slide and fine-grained text embeddings.}
\label{main_fig}
\end{figure*}

Our method consists of six stages. First, the WSI is cropped into patches, and tree-like texts are designed to obtain the data pair. Second, pre-trained encoders are used to extract high-dimensional representations of patches and texts, and a tree-like graph neural network is used as a text prompt encoder to exchange text semantic information. Third, multiple slide-level embeddings are generated through patch-level attention aggregation, with each embedding representing slide features of one node. Fourth, correlations between text prompts and slide-level embeddings are calculated to guide the aggregation of multiple slide-level embeddings to the root according to the branch path, thereby obtaining global WSI features. Fifth, path alignment and tree-aware matching learning loss functions are designed to constrain the relationship of different text semantics. Finally, prediction scores are obtained by calculating the slide-text similarity. Fig.\ref{main_fig} shows the overall process of PathTree.

\subsection{Preparation of WSIs and Tree-like Texts}

\begin{figure}[tbp]
\centering
\includegraphics[width=0.9\linewidth]{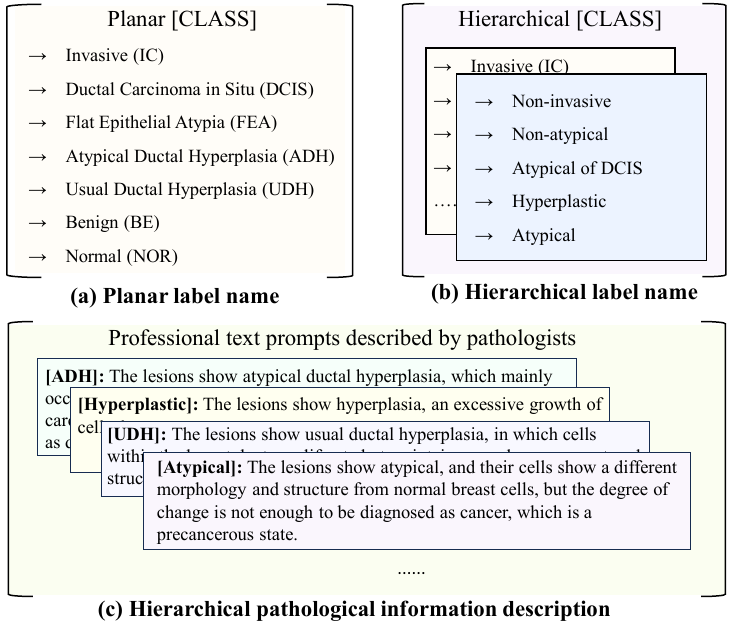}
\caption{Three different text forms of pathological lung tissue categories. (a) Planar text labels, using only fine-grained category names; (b) hierarchical text labels, using both fine-grained and coarse-grained category names; (c) hierarchical descriptive text labels, with each fine-grained and coarse-grained category described in detail using pathology terminology.} 
\label{text_prompt}
\end{figure}

The gigapixel size of WSIs makes it difficult to process the entire image directly. To preprocess WSIs, we use the Otsu method (\citealt{otsu1975threshold}) to separate the foreground in the thumbnail and then perform a sliding window operation at $20\times$ magnification to crop non-overlapping patches with $256 \times 256$ resolutions. Recent contrastive learning (\citealt{radford2021learning}) can obtain well-performing feature extractors by training image-text pairs, therefore, we use the image encoder of PLIP (\citealt{huang2023visual}), which is a model pre-trained on more than 200,000 pathological images paired with natural language descriptions, to extract the high-dimensional embedding of each patch. PathTree prepares text modalities based on specific tasks in histopathology. Conventional text information only contains category words or phrases. For the hierarchical tree structure, there are additional categories on the root and branch nodes. We provide a detailed description of the pathological morphology for each node in the tree structure (including the root, branch, and leaf nodes), which at least two pathology experts verify. Fig.\ref{text_prompt} shows examples of the difference between the conventional planar text, hierarchical text, and the text information we design in lung tissues. Each pathology task has a dedicated tree including hierarchical category descriptions from coarse-grained to fine-grained. Coarse-grained categories (including root nodes) represent the step-by-step judgment method used by pathologists when making diagnoses, and fine-grained categories, i.e. leaf nodes, are the classification results ultimately required for the pathology task. According to the properties of the binary tree (\citealt{li2023deeptree}), we denote the number of fine-grained categories as $N$ and the number of coarse-grained categories as $N-1$.

\subsection{Text Prompt Encoding}

PathTree uses textual semantic information as prompts to guide forward propagation. To match the image encoding, the text encoder of PLIP is utilized to generate tree-like text embeddings $X_{t} \in \mathbb{R}^{(2N-1) \times d}$, where $d$ represents the dimension of embeddings. The content of the prompt text has a significant impact on the downstream performance of contrastive learning models. To avoid excessive time costs on variations in word phrasing, existing research has introduced learnable context tokens to enhance performance (\citealt{zhou2022learning,zhou2022conditional}). However, learnable prefixes and suffixes are often overgeneralized and lack pathological specificity. Therefore, we further design a tree-structure encoder to allow information transfer between text prompts. Specifically, we represent the tree structure as a directed graph $G=\{X_{t}, E_1, E_2\}$, where $E_1$ represents the set of directed edges from parent nodes to child nodes, $E_2$ represents the set of directed edges from child nodes to parent nodes. Their corresponding adjacency matrices are denoted as $A_1$ and $A_2$ respectively. Then, graph attention network (GAT) layers are used to transfer node information dynamically:
\begin{equation}
	H_j^{(1)} = \sigma(\mathrm{GAT}_{j}^{(1)} (X_{t}, A_j)), \text{  } H_j^{(2)} = \sigma(\mathrm{GAT}_{j}^{(2)} (H_j^{(1)}, A_j)),
\end{equation}
where $j=1,2$, $\sigma$ is the activation function, such as ReLU. We use a dual interaction mechanism to deliver the upper and lower-level messages of the tree nodes:
\begin{equation}
	T = \sigma(\mathrm{W_1}(H_1^{(2)} + H_2^{(2)}) + \mathrm{W_2}(H_1^{(2)} \odot H_2^{(2)})),
\end{equation}
where $\mathrm{W_1}$ and $\mathrm{W_2}$ are learnable linear transformation matrices, $T$ represents the text representation with prior knowledge of hierarchical paths.

\subsection{Multiple Attention Module}

\begin{figure}[tbp]
\centering
\includegraphics[width=0.9\linewidth]{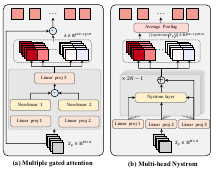}
\caption{Schematic diagram of the two attention modules. (a) Multiple gated attention assigns multiple attention scores to each patch, which are then weighted with the patch embedding to obtain multiple slide representations; (b) multi-head Nystrom assigns $2N-1$ heads, then uses the linear Nystrom method to update the embedding in each head, and finally obtains multiple slide representations through average pooling layer.}
\label{attn_block}
\end{figure}

Two attention modules are introduced to generate multiple slide-level features, as shown in Fig.\ref{attn_block}. The first is multi-gated attention:
\begin{equation}
	I = A \cdot X_p,
\end{equation}
where $I$ is regarded as a matrix composed of the slide-level representation of each node in the tree, and $A$ represents the score matrix,  which is composed of the attention score vectors of all patches and is expressed as follows:
\begin{equation}
	A  = \operatorname{Softmax}( \mathrm{W} (\tanh (\mathrm{U_1} \cdot X_p) \odot \operatorname{sigm}(\mathrm{U_2} \cdot X_p))),
\end{equation}
where $\mathrm{U_1}, \mathrm{U_2} \in \mathbb{R}^{{(d/2) \times d}}$, $\mathrm{W} \in \mathbb{R}^{(2N-1) \times (d/2) }$ are all learnable parameters. 

The second is multi-head Nystrom, where different heads can capture distinct features and patterns of slides in parallel. We set the number of heads to $2N-1$, consistent with the number of nodes in the tree structure. The self-attention mechanism achieves patch interaction by calculating similarity scores, but its standard version leads to high memory and time complexity in long sequence problems like WSIs (\citealt{chu2024retmil}). Therefore, we use the Nystrom method (\citealt{xiong2021nystromformer}) to approximate the self-attention computation to reduce computational overhead. The final embedding of each head is obtained by using the average pooling operation. The form of multi-head self-attention can be defined as follows:
\begin{equation}
	I = [\mathrm{Avg}(\mathrm{nystrom}_{1}(X_p)),\dots, \mathrm{Avg}(\mathrm{nystrom}_{2N-1}(X_p))].
\end{equation}
PathTree generates multiple slide embeddings based on the above methods and then assigns them to the nodes corresponding to the tree structure.

\subsection{Text-guided Tree-like Aggregator}

Text prompt embeddings provide crucial information that helps guide the aggregation of multiple slide-level embeddings along the path of the tree, which is beneficial to generating WSI features that contain hierarchical morphological prior knowledge. Starting from the leaves, we recursively aggregate embeddings of each node from bottom to top until the root. For any node $\gamma$, its two embeddings of child nodes $\alpha$ and $\beta$ are fused as follows:
\begin{equation}
	b_{\gamma} = s_{\alpha}b_{\alpha} + s_{\beta}b_{\beta} + i_{\gamma},
\end{equation}
where
\begin{equation}
	s_{\alpha} = \frac{\exp({b_{\alpha} t_{\alpha}^{\top}})}{\exp(b_{\alpha} t_{\alpha}^{\top}) + \exp(b_{\beta} t_{\beta}^{\top})},\text{ } s_{\beta} = 1-s_{\alpha},
\end{equation}
where $t_{\alpha}$ and $t_{\beta}$ represent the text embeddings corresponding to the two child nodes. $b_{\gamma}$ is the slide-level embedding of node $\gamma$ after fusion. Note that when a child node is a leaf, its original embedding $i_{\gamma}$ is used as the input $b_{\gamma}$. Algorithm \ref{tree_agg} describes the entire recursive aggregation process under programming logic. When the aggregation is complete, the root integrates the information of all nodes, and its slide-level embedding serves as the global WSI feature.

\begin{algorithm}[H]
\caption{Text-guided Tree-like Aggregator}
\label{tree_agg}
\begin{algorithmic}[1]
\renewcommand{\algorithmicrequire}{\textbf{function}}
\renewcommand{\algorithmicensure}{\textbf{end function}}
\REQUIRE $\mathrm{SlideAggregator}$($tree$, $T$, $I$) 
\STATE $\rhd$$Input$ $tree$: recursive tree node dictionary, each $node$ includes $id$ and $child$. \\
(if $node$ is a leaf, $node['child'] = None$).
\STATE $\rhd$$Input$ $T$: $[t_1, t_2, \dots, t_{2N-1}]^{\top}$, text prompt embeddings.
\STATE $\rhd$$Input$ $I$: $[i_1, i_2, \dots, i_{2N-1}]^{\top}$, multiple slide embeddings. 
\STATE $S \gets I \cdot T^{\top}$      \hspace{1.5cm}$\rhd$ Compute a contrast matrix $S$. 
\STATE $D:=[d_1, d_2, \dots, d_{2N-1}] \gets diag(S)$ 
\STATE  \hspace{4.2cm}$\rhd$ Get the diagonal of $S$. 
\STATE $F \gets \{ \}$ \\
    \hspace{0.6cm}$\rhd$ Create an empty dictionary to save embeddings.
\STATE $F \gets \mathrm{FuseBranch}\space(tree, I, D, F)$ 
\STATE  \hspace{2.cm}$\rhd$ Recursively fuse all slide embeddings.
\STATE $g \gets F[tree['id']]$ \hspace{1.8cm} $\rhd$ Global WSI feature.
\RETURN $g$
\ENSURE 
\STATE
\REQUIRE $\mathrm{FuseBranch}$($node$, $I$, $D$, $F$) 
    \IF {\textbf{not} $child$ in $node$}
        \STATE $F[node['id']] \gets i_{node['id']}$ 
        \STATE \hspace{0.8cm} $\rhd$ Assign the slide embedding of the leaf to $F$.
    \ELSE 
        \FOR {$child$ in $node['child']$}
            \STATE $F \gets \mathrm{FuseBranch}(child, I, D, F)$ 
            \STATE	\hspace{1.1cm}$\rhd$ Recursively aggregate child information.
        \ENDFOR
        \STATE $sibling \gets \mathrm{FindSibling}(tree, child['id'])$ 
        \STATE	\hspace{4.cm}$\rhd$ Find its sibling node.
        \STATE $[s_{\alpha},s_{\beta}] \gets \mathrm{Softmax}(d_{child['id']}, d_{sibling['id]})$ 
        \STATE  \hspace{4.2cm}$\rhd$ Information weight.
        \STATE $[b_{\alpha}, b_{\beta}] \gets [F[child['id']], F[sibling['id']]]$
        \STATE $b_{\gamma} \gets [s_{\alpha},s_{\beta}] \cdot [b_{\alpha}, b_{\beta}]^{\top}$ \hspace{0.35cm}$\rhd$ Fuse child embeddings.
        \STATE $F[node['id']] \gets b_{\gamma} + i_{node['id']}$
        \STATE \hspace{4.1cm}$\rhd$ Update embeddings.
    \ENDIF
\RETURN $F$
\ENSURE
\end{algorithmic}
\end{algorithm}

\subsection{Joint Slide-text Constraints}

\begin{figure}[tbp]
	\centering
	\includegraphics[width=0.45\linewidth]{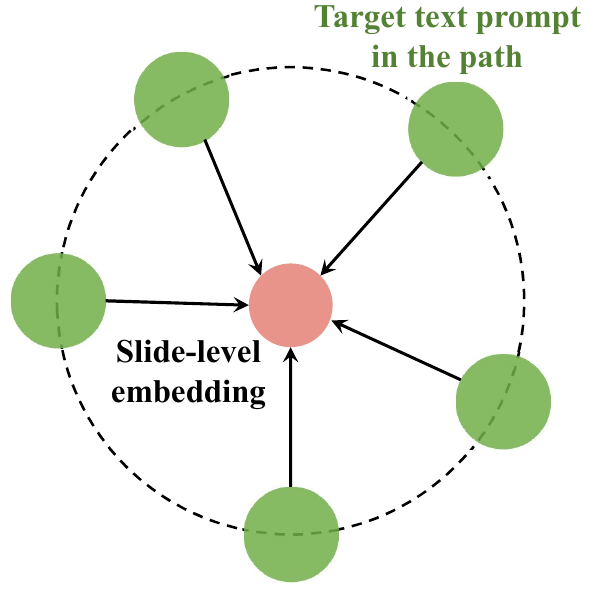}
	\caption{Graphical demonstration of path alignment learning. Its goal is to make the slide-level embedding of the target node close to all text embeddings in its path to the root node.}
    \label{loss_path}
	\end{figure}

\textbf{Path Alignment Learning}: The hierarchical classification requires the WSI feature to align with the corresponding fine-grained and coarse-grained categories. Inspired by research on hierarchical text classification (\citealt{zhou2020hierarchy,chen2021hierarchy}), our proposed PathTree aligns slide-level representation with text prompt embeddings from the root to the corresponding leaf to achieve this goal. Fig.\ref{loss_path} demonstrates the schematic diagram of path alignment learning. Specifically, for a global WSI feature $g$ and its fine-grained text prompt embedding, the mean square error metric loss is designed to constrain their alignment degree:
\begin{equation}
	\mathcal{L}_{path} = \frac{1}{|P|} \sum_{k \in P}^{} \left \| g - t_k \right \|_2^2,
\end{equation}
where $P$ represents the set of all nodes in the path starting from the root and ending at the corresponding leaf. By minimizing between the slide-level representation and the embeddings of the fine and coarse-grained text prompts under the tree structure path, $\mathcal{L}_{path}$ helps the WSI feature to fit all hierarchical labels better and prevent from being alienated from the tree structure information driven by the entire problem, causing weak generalization performance. PathTree uses the depth-first search (DFS) (\citealt{tarjan1972depth}) to find the path of the current node, and Algorithm \ref{findpath} shows its pseudocode.

\begin{algorithm}[H]
	\caption{Finding Paths with DFS Algorithm}
	\label{findpath}
	\begin{algorithmic}[1]
	\renewcommand{\algorithmicrequire}{\textbf{function}}
	\renewcommand{\algorithmicensure}{\textbf{end function}}
	\REQUIRE $\mathrm{FindPath}$($node$, $target['id']$, $path$) 
	\STATE $\rhd$$Input$ $path$: $[\text{ }]$, create it to save $id$. 
	\STATE $path \gets \mathrm{Append}(path, node)$
	\IF {$node['id'] = target['id']$}
		\RETURN $path$
	\ENDIF
	\IF {$child$ in $node$}
		\FOR {$child$ in $node['child']$} 
			\STATE $Path \gets \mathrm{Copy}(Path)$
			\STATE $result \gets \mathrm{FindPath}(child, target['id'], Path)$ 
		\IF {$result$ \textbf{is not} $None$}
			\RETURN $results$
		\ENDIF
		\ENDFOR
	\ENDIF
	\ENSURE
	\end{algorithmic}
\end{algorithm}

\begin{figure}[tbp]
	\centering
	\includegraphics[width=1\linewidth]{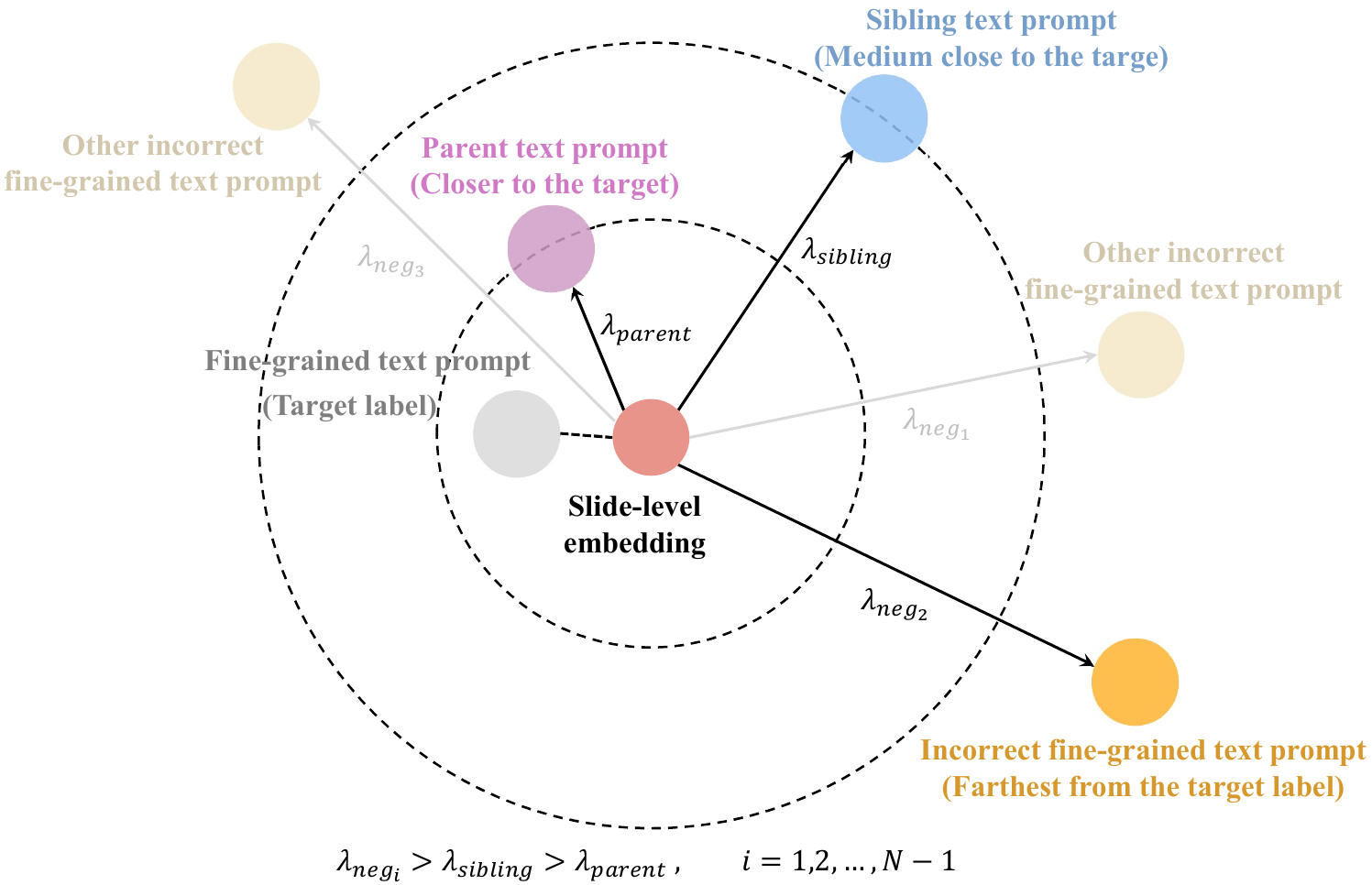}
	\caption{Graphical demonstration of tree-aware matching learning, where the slide-level embedding of the target node is used as the anchor, its text embedding as the positive sample, and other text embeddings as the negative sample. The goal is to make the distance between the anchor and the negative sample larger than the distance to the positive sample.}
    \label{loss_match}
	\end{figure}

\textbf{Tree-aware Matching Learning}: In addition to aligning with the fine-grained text prompt, the WSI features extracted by PathTree are expected to match three other text prompts in the tree structure. Specifically, the parent node (containing the coarse-grained text prompt) should produce strong correlations with WSI features, sibling nodes should produce secondary correlations, and other incorrect fine-grained text prompts should stay away. We introduce the triplet loss $\mathcal{L}_{neg}$ to model the above relationships. $\mathcal{L}_{neg}$ aims to use a margin $\lambda_{neg}$ to penalize the distance between the slide representation and the wrong text semantics, which is expressed as follows:
\begin{equation}
	\mathcal{L}_{neg} = \mathrm{max}(0, D(g, t_{pos}) - D(g, t_{neg}) + \lambda_{neg}),
\end{equation}
where $D$ represents the metric function based on the $L_2$ norm, $t_{pos}$ and $t_{neg}$ represent correct and incorrect prompt embeddings, $\lambda$ is a marginal constant, the larger it is, the greater the penalty. We denote parent, sibling, and other leaf nodes as negative nodes, and set $\lambda_{leaf} > \lambda_{sibling} > \lambda_{parent}$. We take the sum of all the triplet losses as the final $\mathcal{L}_{match}$, which is expressed as follows:
\begin{equation}
	\mathcal{L}_{match} = \mathcal{L}_{parent} + \mathcal{L}_{sibling} + \textstyle \frac{1}{N-1}\textstyle \sum \mathcal{L}_{leaf}.
\end{equation}
Fig.\ref{loss_match} demonstrates the schematic diagram of tree-aware matching learning.


\subsection{Slide Prediction with Text Prompts}

Unlike conventional single-modality WSI classification methods that directly utilize labels as integers for training, PathTree obtains prediction scores by calculating the similarity between WSI features and text prompt embeddings, which improves flexibility in how the model predicts and increases few-shot learning capabilities for downstream tasks. We use text prompt embeddings $T = [t_1, t_2, \dots, t_N]^{\top}$ as the classification weight vector to calculate predicted probabilities:
\begin{equation}
	p(y=j|I) = \frac{\exp(\mathrm{sim}(t_j, g)/ \tau)}{ {\textstyle \sum_{n=1}^{N} \exp(\mathrm{sim}(t_n, g)/ \tau)} },
\end{equation}
where $\tau$ is a learnable temperature parameter, $\mathrm{sim(\cdot)}$ represents cosine similarity. The probability is finally fed into cross-entropy loss $\mathcal{L}_{ce}$ to minimize the standard classification loss. The entire loss of PathTree is as follows:
\begin{equation}
	\mathcal{L}_{all} = \mathcal{L}_{ce} + \mu_{m}\mathcal{L}_{match}+ \mu_{p}\mathcal{L}_{path},
\end{equation}
where $\mu_{m}$ and $\mu_{p}$ are the coefficients of the two joint constraint losses. In summary, PathTree is trained through $\mathcal{L}_{path}$, $\mathcal{L}_{match}$, and the target domain loss corresponding to the downstream task, allowing WSI features to perform constrained exploration in the tree-like text feature space and contextual learning.

\section{Experiments}
\subsection{Datasets}

\textbf{In-house SYSFL:} We collect 1189 digital intraoperative cryosections of lung tissue from the First Affiliated Hospital of Sun Yat-sen University to construct the SYSFL dataset, which includes seven categories with atypical adenomatous hyperplasia (AAH), adenocarcinoma in situ (AIS), minimally invasive adenocarcinoma (MIA), invasive adenocarcinoma (IAC), squamous cell carcinoma (SCC), pneumonia (PNE) and normal lung tissue (NOR). They are all acquired using the Teksqray slide system and scanned with $40\times$ magnification. Note that AAH, AIS, MIA, and IAC present progressive symptoms and are difficult to diagnose on cryosections. All slide-level labels are based on the description of the frozen pathological report and verified by two professional pathologists. Due to the limitations of preparation methods and diagnostic time, diagnosing lung tissue cryosections is more difficult than conventional paraffin-embedded pathological sections. Computer-aided diagnosis can help pathologists reduce the burden of reviewing cryosections of lung tissue. 

\begin{figure}[tbp]
\centering
\includegraphics[width=0.8\linewidth]{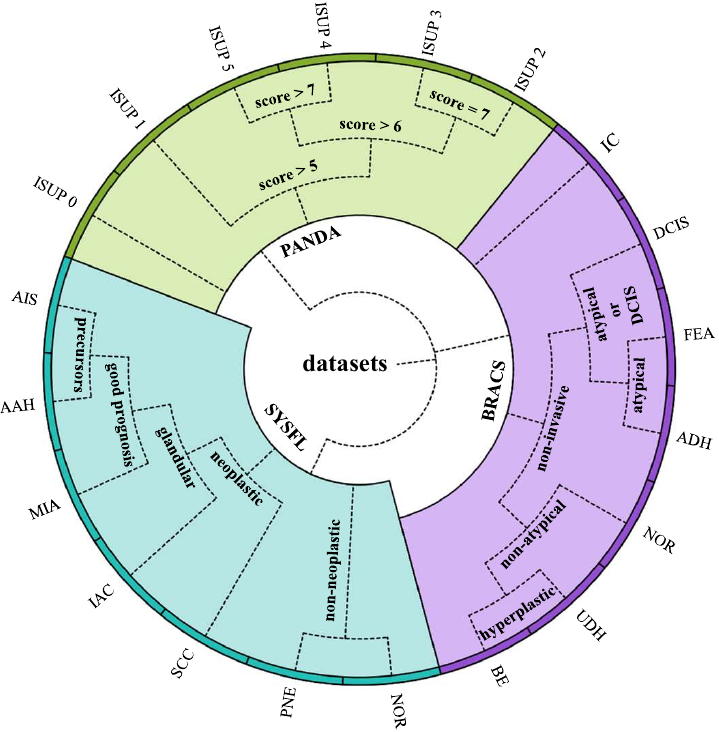}
\caption{The hierarchical forms of SYSFL, PANDA, and BRACS datasets. They are all represented by binary trees.}
\label{dataset_circle}
\end{figure}

\begin{table}[tbp]
\caption{Comparison results of fine-grained classification}
\label{datasets_num}
\centering
\resizebox{1.\columnwidth}{!}{
\renewcommand{\arraystretch}{1.5}{
\begin{tabular}{clllllll}
\hline
\multicolumn{1}{l}{Datasets} & \multicolumn{7}{c}{The number of WSIs per fine-grained class} \\ \hline
\multirow{2}{*}{SYSFL}       & AAH    & AIS    & MIA    & IAC    & SCC   & PNE   & NOR                     \\
                             & 102  & 354  & 137  & 244  & 111 & 153 & 86                   \\ \hline
\multirow{2}{*}{PANDA}       & ISUP 0    & ISUP 1    & ISUP 2    & ISUP 3    & ISUP 4   & ISUP 5   & \multicolumn{1}{c}{-} \\
                             & 2892  & 2666  & 1343  & 1242  & 1249 & 1224 & \multicolumn{1}{c}{-} \\ \hline
\multirow{2}{*}{BRACS}       & IC    & DCIS    & FEA    & ADH    & NOR   & UDH   & BE                     \\
                             & 132  & 61  & 41  & 48  & 44 & 74 & 147                   \\ \hline
\end{tabular}}}
\end{table}

\begin{table*}[htbp]
\caption{ACC, AUC, and Weighted F1 results [\%] on the fine-grained 7-category SYSFL, 6-category PANDA, and 7-category BRACS datasets. Best performance in \textbf{bold}, second best \underline{underlined}}
\label{fine_planar}
\centering
\resizebox{1.73\columnwidth}{!}{
\renewcommand{\arraystretch}{1.3}{
\begin{tabular}{lccccccccc}
\toprule
\multicolumn{1}{l}{\multirow{2}{*}{Methods}} &
  \multicolumn{3}{c}{SYSFL} &
  \multicolumn{3}{c}{PANDA} &
  \multicolumn{3}{c}{BRACS} \\ \cmidrule(lr){2-4} \cmidrule(lr){5-7} \cmidrule(lr){8-10} 
\multicolumn{1}{c}{} &
  ACC &
  AUC &
  Weighted F1 &
  ACC &
  AUC &
  Weighted F1 &
  ACC &
  AUC &
  Weighted F1 \\ \midrule
ABMIL (\citealt{ilse2018attention}) &
  $59.55_{7.01}$ & 
  $87.36_{2.29}$ &
  \multicolumn{1}{c|}{$56.69_{4.83}$} &
  $58.55_{0.54}$ &
  $85.19_{0.73}$ &
  \multicolumn{1}{c|}{$57.09_{0.43}$} &
  $51.18_{3.00}$ &
  $77.03_{2.62}$ &
  $43.82_{3.37}$ \\
CLAM (\citealt{lu2021data})&
  $58.46_{7.19}$ &  
  $85.52_{3.06}$ &
  \multicolumn{1}{c|}{$57.58_{5.83}$} &
  $60.29_{0.88}$ &
  $\underline{87.48_{0.51}}$ &
  \multicolumn{1}{c|}{$59.51_{0.57}$} &
  $53.57_{3.20}$ &
  $78.22_{4.86}$ &
  $51.55_{2.17}$ \\
DSMIL (\citealt{li2021dual})&
  $59.55_{5.28}$ & 
  $87.48_{2.10}$ &
  \multicolumn{1}{c|}{$58.39_{3.70}$} & 
  $59.96_{0.45}$ & 
  $87.08_{0.42}$ & 
  \multicolumn{1}{c|}{$58.82_{0.54}$} &
  $53.94_{5.19}$ & 
  $80.33_{3.71}$ & 
  $52.47_{4.54}$ \\
TransMIL (\citealt{shao2021transmil})&
  $55.42_{7.16}$ &  
  $84.19_{4.27}$ & 
  \multicolumn{1}{c|}{$54.01_{5.66}$} &
  $59.45_{1.84}$ & 
  $86.65_{1.15}$ & 
  \multicolumn{1}{c|}{$58.00_{2.61}$} & 
  $50.09_{1.80}$ & 
  $76.15_{2.74}$ & 
  $48.35_{1.79}$ \\ 
GTP (\citealt{zheng2022graph})&
  $46.93_{10.60}$&
  $74.77_{8.07}$ &          
  \multicolumn{1}{c|}{$44.98_{9.30}$} &
  $50.45_{2.38}$ & 
  $80.93_{1.13}$ & 
  \multicolumn{1}{c|}{$48.86_{2.32}$} &  
  $41.33_{8.14}$ &  
  $71.29_{9.01}$ & 
  $39.97_{9.43}$ \\ 
DTFD-MIL (\citealt{zhang2022dtfd})&
  $53.40_{4.90}$ &  
  $84.73_{2.00}$ & 
  \multicolumn{1}{c|}{$51.64_{3.66}$} &
  $55.92_{0.62}$ &
  $84.58_{0.70}$ & 
  \multicolumn{1}{c|}{$54.84_{0.69}$} & 
  $\underline{53.57_{1.69}}$ & 
  $79.25_{3.39}$ &  
  $52.26_{2.03}$ \\
  HIPT (\citealt{chen2022scaling})&
  $51.05_{3.95}$ &
  $77.66_{2.20}$ &
  \multicolumn{1}{c|}{$47.00_{4.41}$} & 
  $57.22_{0.68}$ &
  $83.92_{0.63}$ &
  \multicolumn{1}{c|}{$54.90_{1.42}$} &
  $48.16_{2.00}$ &
  $76.29_{2.88}$ &
  $48.55_{2.51}$ \\
  ILRA (\citealt{xiang2022exploring})&
  $60.39_{5.51}$ & 
  $86.32_{2.16}$ & 
  \multicolumn{1}{c|}{$58.66_{3.17}$} & 
  $60.41_{1.05}$ & 
  $86.43_{0.99}$ & 
  \multicolumn{1}{c|}{$59.73_{0.76}$} & 
  $51.74_{3.82}$ & 
  $77.33_{6.55}$ & 
  $50.45_{2.66}$ \\  \hline
\textbf{PathTree (Ours)} &
  $\underline{61.75_{4.32}}$ & 
  $\underline{88.49_{2.21}}$ & 
  \multicolumn{1}{c|}{$\underline{59.48_{3.52}}$} & 
  $\underline{61.14_{0.96}}$ & 
  $87.39_{0.47}$ & 
  \multicolumn{1}{c|}{$\underline{60.27_{1.23}}$} &
  $\mathbf{55.22_{4.62}}$ & 
  $\mathbf{81.81_{2.43}}$ &
  $\mathbf{54.64_{4.11}}$ \\
\textbf{PathTree-Ny (Ours)} &
  $\mathbf{63.68_{4.87}}$ & 
  $\mathbf{88.93_{0.67}}$ & 
  \multicolumn{1}{c|}{$\mathbf{62.04_{3.74}}$} & 
  $\mathbf{62.35_{0.65}}$ & 
  $\mathbf{88.26_{0.33}}$ & 
  \multicolumn{1}{c|}{$\mathbf{61.54_{0.51}}$} &
  $51.71_{1.72}$ & 
  $\underline{80.34_{2.41}}$ &
  $\underline{52.61_{1.91}}$ \\  \bottomrule 
\end{tabular}}}
\end{table*}

\begin{table*}[htbp]
\caption{H-Precision, H-Recall, and H-F1 results [\%] on the fine-grained 7-category SYSFL, 6-category PANDA, and 7-category BRACS datasets.}
\label{fine_hier}
\centering
\resizebox{1.73\columnwidth}{!}{
\renewcommand{\arraystretch}{1.3}{
\begin{tabular}{lccccccccc}
\toprule
\multicolumn{1}{l}{\multirow{2}{*}{Methods}} &
  \multicolumn{3}{c}{SYSFL} &
  \multicolumn{3}{c}{PANDA} &
  \multicolumn{3}{c}{BRACS} \\ \cmidrule(lr){2-4} \cmidrule(lr){5-7} \cmidrule(lr){8-10} 
\multicolumn{1}{c}{} &
  H-Precision &
  H-Recall &
  H-F1 &
  H-Precision &
  H-Recall &
  H-F1 &
  H-Precision &
  H-Recall &
  H-F1    \\ \midrule 
ABMIL (\citealt{ilse2018attention}) &
  $76.52_{4.78}$ & 
  $80.36_{5.61}$ & 
  \multicolumn{1}{c|}{$78.37_{4.97}$} & 
  ${70.91_{0.74}}$ & 
  $71.22_{0.68}$ & 
  \multicolumn{1}{c|}{$71.06_{0.61}$} & 
  $65.33_{2.21}$ & 
  $66.51_{2.42}$ & 
  $65.90_{2.06}$ \\ 
CLAM (\citealt{lu2021data})&
  $77.51_{3.56}$ &  
  $78.26_{3.80}$ & 
  \multicolumn{1}{c|}{$77.86_{3.37}$} & 
  $73.59_{1.70}$ &
  $72.92_{1.72}$ & 
  \multicolumn{1}{c|}{$73.22_{0.37}$} & 
  $66.06_{4.71}$ & 
  $67.70_{4.95}$ & 
  $66.86_{4.79}$ \\ 
DSMIL (\citealt{li2021dual})&
  $77.38_{3.67}$ &  
  $80.37_{3.14}$ &  
  \multicolumn{1}{c|}{$78.84_{3.24}$} &  
  $73.71_{0.70}$ & 
  $71.53_{0.51}$ & 
  \multicolumn{1}{c|}{$72.60_{0.31}$} & 
  $68.34_{2.82}$ & 
  $\underline{71.19_{4.03}}$ & 
  $\underline{69.72_{3.30}}$ \\ 
TransMIL (\citealt{shao2021transmil})& 
  $73.54_{6.94}$ &  
  $76.66_{4.27}$ & 
  \multicolumn{1}{c|}{$74.98_{5.16}$} & 
  $73.06_{1.30}$ & 
  $71.56_{1.92}$ & 
  \multicolumn{1}{c|}{$72.30_{1.43}$} & 
  $66.51_{1.28}$ & 
  $65.97_{2.43}$ & 
  $66.22_{1.37}$ \\ 
GTP (\citealt{zheng2022graph})&
  $71.83_{6.50}$ & 
  $70.85_{4.96}$ &
  \multicolumn{1}{c|}{$71.25_{5.11}$} &
  $63.59_{3.37}$ & 
  $66.74_{0.69}$ & 
  \multicolumn{1}{c|}{$65.09_{1.79}$} &  
  $57.98_{9.22}$ &  
  $59.74_{9.04}$ & 
  $58.75_{8.78}$ \\  
DTFD-MIL (\citealt{li2021dual})&
  $70.96_{6.69}$ &  
  $78.82_{2.76}$ &
  \multicolumn{1}{c|}{$74.62_{4.81}$} & 
  $68.82_{0.75}$ & 
  $69.62_{0.73}$ & 
  \multicolumn{1}{c|}{$69.21_{0.54}$} & 
  $\mathbf{70.33_{1.73}}$ & 
  $69.06_{2.25}$ &  
  $69.67_{1.61}$ \\ 
  HIPT (\citealt{chen2022scaling})&
  $71.34_{3.80}$ &
  $71.26_{5.22}$ & 
  \multicolumn{1}{c|}{$71.22_{3.58}$} &
   $\underline{73.85_{0.46}}$ &
   $68.67_{1.06}$ &
  \multicolumn{1}{c|}{$71.16_{0.67}$} &
   $65.52_{2.29}$ &
   $67.63_{1.11}$ &
   $66.56_{1.71}$ \\
  ILRA (\citealt{xiang2022exploring})&
  $77.31_{5.27}$ & 
  $\mathbf{81.23_{5.08}}$ & 
  \multicolumn{1}{c|}{$79.22_{5.14}$} & 
  $73.12_{1.67}$ & 
  $72.49_{0.69}$ & 
  \multicolumn{1}{c|}{$72.79_{0.85}$} & 
  $67.15_{2.92}$ & 
  $70.36_{1.95}$ & 
  $68.71_{2.37}$ \\ \hline
\textbf{PathTree (Ours)} &
  $\underline{79.21_{3.79}}$ & 
  $\underline{80.91_{3.12}}$ & 
  \multicolumn{1}{c|}{$\underline{80.03_{3.09}}$} & 
  $73.24_{1.63}$ & 
  $\underline{73.38_{1.07}}$ & 
  \multicolumn{1}{c|}{$\underline{73.29_{0.50}}$} &
  $\underline{70.20_{4.64}}$ &
  $\mathbf{72.03_{3.47}}$ & 
  $\mathbf{71.09_{3.97}}$ \\
\textbf{PathTree-Ny (Ours)} &
  $\mathbf{80.62_{3.55}}$ & 
  $80.79_{2.75}$ & 
  \multicolumn{1}{c|}{$\mathbf{80.69_{2.96}}$} & 
  $\mathbf{74.31_{0.78}}$ & 
  $\mathbf{74.16_{1.89}}$ & 
  \multicolumn{1}{c|}{$\mathbf{74.22_{0.56}}$} &
  $66.85_{2.36}$ &
  $70.42_{1.77}$ & 
  $68.58_{1.86}$ \\ \bottomrule 
\end{tabular}}}
\end{table*}

\textbf{Public PANDA (\citealt{bulten2022artificial}):} This large-scale prostate dataset contains six categories, including 10618 H\&E-stained biopsy WSIs. Based on the ISUP grading index, the cancer level can be divided into 0 (no cancer), 1 (Gleason score: $3+3=6$), 2 (Gleason score: $3+4=7$), 3 (Gleason score: $4+3=7$), 4 (Gleason score: $4+4=8; 3+5=8; 5+3=8$) and 5 (Gleason score: $4+5=9$; $5+5=10$). Due to the complexity of the grading system, it is challenging to classify WSI directly through slice-level labels.

\textbf{Public BRACS (\citealt{brancati2022bracs}):} This breast cancer dataset contains 547 H\&E-stained WSIs collected from 189 patients, including seven categories with normal breast tissue (NOR), benign lesion (BE), usual ductal hyperplasia (UDH), flat epithelial atypia (FEA), atypical ductal hyperplasia (ADH), ductal carcinoma in situ (DCIS) and invasive ductal and lobular carcinoma (IC). Due to the diversity of breast cancer subtypes, pathologists can classify them in a tree-like progressive manner, where the classification difficulty increases with each level. 

According to actual diagnostic patterns, three datasets are refined into hierarchical structures certified by authoritative pathology experts to simulate real diagnostic scenarios. These provide resources for evaluating PathTree and related WSI analysis methods. The hierarchical decision trees for coarse-to-fine classification of the three datasets are shown in Figure \ref{dataset_circle}. The specific details are shown in Table \ref{datasets_num}.

\begin{table*}[tbp]
\caption{ACC, AUC, and weighted F1 results [\%] on the coarse-grained 4-category SYSFL, 2-category PANDA, and 3-category BRACS datasets.}
\label{coarse_label}
\centering
\resizebox{1.73\columnwidth}{!}{
\renewcommand{\arraystretch}{1.3}{
\begin{tabular}{lccccccccc}
\toprule
\multicolumn{1}{l}{\multirow{2}{*}{Methods}} &
  \multicolumn{3}{c}{SYSFL} &
  \multicolumn{3}{c}{PANDA} &
  \multicolumn{3}{c}{BRACS} \\ \cmidrule(lr){2-4} \cmidrule(lr){5-7} \cmidrule(lr){8-10} 
\multicolumn{1}{c}{} &
  ACC &
  AUC &
  Weighted F1 &
  ACC &
  AUC &
  Weighted F1 &
  ACC &
  AUC &
  Weighted F1 \\ \midrule
ABMIL (\citealt{ilse2018attention}) &
  $77.83_{6.82}$ &  
  $92.40_{3.45}$ & 
  \multicolumn{1}{c|}{$76.90_{7.32}$} & 
  $87.66_{0.95}$ & 
  $93.07_{0.47}$ & 
  \multicolumn{1}{c|}{$87.74_{0.93}$} &
  $69.83_{3.26}$ & 
  $81.25_{3.62}$ & 
  $64.88_{4.07}$ \\ 
CLAM (\citealt{lu2021data})&
  $76.99_{4.53}$ &  
  $91.44_{2.60}$ &
  \multicolumn{1}{c|}{$76.88_{4.61}$} &
  $87.81_{0.60}$ & 
  $\underline{94.16_{0.48}}$ &
  \multicolumn{1}{c|}{$88.02_{0.45}$} & 
  $70.58_{4.68}$ & 
  $83.28_{5.86}$ &
  $70.31_{5.13}$ \\
DSMIL (\citealt{li2021dual})&
  $78.93_{5.05}$ &  
  $92.49_{2.01}$ & 
  \multicolumn{1}{c|}{$78.24_{5.82}$} &
  $87.03_{0.90}$ & 
  $93.68_{0.53}$ &  
  \multicolumn{1}{c|}{$87.18_{1.03}$} & 
  $\mathbf{75.32_{4.08}}$ & 
  $\underline{84.85_{3.75}}$ & 
  $\underline{74.27_{3.62}}$ \\
TransMIL (\citealt{shao2021transmil})&
  $72.02_{7.92}$ & 
  $89.11_{5.02}$ & 
  \multicolumn{1}{c|}{$71.36_{8.61}$} & 
  $87.31_{0.50}$ &
  $93.75_{0.40}$ &
  \multicolumn{1}{c|}{$87.50_{0.40}$} & 
  $69.11_{2.05}$ & 
  $80.99_{2.67}$ & 
  $67.93_{3.08}$ \\ 
GTP (\citealt{zheng2022graph})& 
  $66.32_{7.40}$& 
  $80.95_{3.29}$ &      
  \multicolumn{1}{c|}{$65.91_{9.87}$} & 
  $83.07_{2.41}$ & 
  $90.28_{1.23}$ & 
  \multicolumn{1}{c|}{$82.95_{2.72}$} &  
  $65.46_{10.57}$ &  
  $76.63_{9.33}$ & 
  $63.14_{11.05}$ \\ 
DTFD-MIL (\citealt{zhang2022dtfd})&
  $71.68_{6.89}$ &  
  $88.36_{3.77}$ & 
  \multicolumn{1}{c|}{$69.65_{9.01}$} & 
  $84.90_{0.71}$ & 
  $91.52_{0.28}$ & 
  \multicolumn{1}{c|}{$84.89_{0.72}$} & 
  $72.76_{3.53}$ & 
  $83.49_{5.69}$ &  
  $72.13_{3.65}$ \\
  HIPT (\citealt{chen2022scaling})&
  $67.16_{2.88}$ & 
  $85.32_{2.62}$ &
  \multicolumn{1}{c|}{$62.59_{6.88}$} & 
  $86.24_{0.83}$ & 
  $91.61_{0.58}$ & 
  \multicolumn{1}{c|}{$86.62_{0.87}$} & 
  $70.26_{3.34}$ & 
  $80.90_{3.45}$ &
  $70.08_{2.83}$ \\ 
  ILRA (\citealt{xiang2022exploring})&
  $\underline{78.93_{4.99}}$ & 
  $91.23_{1.43}$ & 
  \multicolumn{1}{c|}{$77.64_{5.71}$} &
  $87.16_{0.31}$ & 
  $93.50_{0.54}$ & 
  \multicolumn{1}{c|}{$87.12_{0.42}$} &  
  $74.59_{2.98}$ & 
  $83.70_{6.31}$ &
  $73.44_{3.76}$ \\  \hline 
\textbf{PathTree (Ours)} & 
  $\mathbf{80.11_{4.76}}$ &  
  $\mathbf{93.90_{1.66}}$ & 
  \multicolumn{1}{c|}{$\mathbf{79.46_{4.76}}$} & 
  $\mathbf{88.11_{0.43}}$ & 
  $\mathbf{94.31_{0.26}}$ & 
  \multicolumn{1}{c|}{$\mathbf{88.18_{0.41}}$} & 
  $\underline{74.60_{4.28}}$ & 
  $\mathbf{86.00_{3.09}}$ & 
  $\mathbf{74.28_{4.28}}$ \\
\textbf{PathTree-Ny (Ours)} & 
  $78.59_{4.52}$ &  
  $\underline{93.16_{2.07}}$ & 
  \multicolumn{1}{c|}{$\underline{78.61_{4.46}}$} & 
  $\underline{88.07_{0.78}}$ & 
  $94.06_{0.65}$ & 
  \multicolumn{1}{c|}{$\underline{88.17_{0.71}}$} & 
  $70.20_{1.83}$ & 
  $85.59_{2.71}$ & 
  $71.58_{2.16}$ \\  \bottomrule  
\end{tabular}}}
\end{table*}

\subsection{Baselines}

We consider eight major WSI analysis baselines for direct comparison: ABMIL (\citealt{ilse2018attention}), CLAM (\citealt{lu2021data}), DSMIL (\citealt{li2021dual}), TransMIL (\citealt{shao2021transmil}), GTP (\citealt{zheng2022graph}), DTFD-MIL (\citealt{zhang2022dtfd}), HIPT (\citealt{chen2022scaling}), ILRA (\citealt{xiang2022exploring}), which are the methods that have been widely discussed recently. In addition, three baselines are used for comparison to explore the few-shot performance of our PathTree: (1) Linear-Probe (attention pooling) aggregates all patch features through a learnable attention pooling layer and uses a linear layer to classify the aggregated global features. (2)  CoOp (\citealt{zhou2022learning}) concatenates additional learnable tokens to the text prompt based on Linear-Probe, and then performs classification based on feature alignment between text and slide. (3) TOP (\citealt{qu2024rise}) generates text prototypes of multiple pathology prompt groups through GPT-4 (\citealt{achiam2023gpt}), then designs a prompt-guided pooling layer for aggregating patch features into bag features, and finally performs classification through text-slide feature alignment.

\subsection{Evaluation Metrics}

To comprehensively evaluate the pathological classification task of the tree-like structure, planar and hierarchical metrics are adopted. For planar metrics, we report accuracy (ACC), AUC, and Weighted F1-score to evaluate the fine-grained performance. For hierarchical metrics, each WSI has a multi-label from coarse-grained to fine-grained levels, we additionally report hierarchical precision (H-Precision), recall (H-Recall), and F1-score (H-F1), which are expressed as follows:
\begin{equation}
    \textrm{H-Precision} = \frac{ {\textstyle \sum_{j} |C_t(j) \cap C_p(j)|} }{ {\textstyle \sum_{j} |C_t(j)|} },
\end{equation}
\begin{equation}
    \textrm{H-Recall} = \frac{ {\textstyle \sum_{j} |C_t(j) \cap C_p(j)|} }{ {\textstyle \sum_{j} |C_t(j)|} },
\end{equation}
\begin{equation}
    \textrm{H-F1} = 2\frac{\textrm{H-Precision} \cdot \textrm{H-Recall}}{\textrm{H-Precision} + \textrm{H-Recall}},
\end{equation}
where $C_t(j)$ and $C_p(j)$ are the true and predicted hierarchical label sets of the $j$th sample. Note that the root node is not included in the label set. By calculating the number of intersections between the true and predicted node sets, three metrics can be quantified to show the differences in hierarchical label predictions between models. All results for the three datasets are obtained using 5-fold cross-validation, and each metric is reported as a percentage value and standard error. 

\subsection{Implementation}
In the training phase, we employ the Adam (\citealt{kingma2014adam}) optimizer, with a learning rate of $3\times 10^{-4}$, betas of 0.9 to 0.98, and weight decay of $10^{-4}$. The epoch is set to 100, and the batch size is set to 1, meaning only one WSI is processed in each iteration. For the slide-text joint loss function, we set $\lambda_{leaf},\lambda_{sibling},\lambda_{parent}$ to $0.2$, $0.1$, and $0.002$ respectively; for slide-text similarity, we initialize $\tau$ to 0.07; for $\mu_{m}$ and $\mu_{p}$, we set them to 1 in the comparative experiments with other baselines, and explore their impact in the ablation experiments. We use the image encoder of PLIP for each baseline and PathTree, as a fair comparison, and all experiments are conducted with the same hyperparameters and data partitioning settings.

\subsection{Results and Analysis}

\subsubsection{Fine-grained Classification Results}

Table \ref{fine_planar} presents the planar metric results for fine-grained classification tasks on the SYSFL, PANDA, and BRACS datasets. Overall, the PathTree based on gated attention outperforms other competitors in all three metrics, and the PathTree-Ny based on multi-head Nystrom mechanisms achieves even higher performance on the SYSFL and PANDA datasets compared to PathTree. For instance, in the weighted F1 metric on SYSFL, PathTree-Ny surpasses PathTree by 2.56\% and the best alternative WSI analysis method by 3.38\%. In the PANDA dataset, the accuracy of PathTree-Ny exceeds that of PathTree by approximately 1.21\%, equating to a difference of nearly 128 WSIs. However, in the BRACS dataset, PathTree-Ny performs less effectively than PathTree, possibly due to differences in the classification tasks. In differentiating breast cancer subtypes, a single slide may contain multiple regions related to other categories. Pathologists determine the category of a slide based on the region with the highest malignancy. However, such regions are often small, making it difficult for self-attention computation to focus on these regions, leading to incorrect judgments. Conversely, pathologists focus more on global regions in tasks such as distinguishing lung tissue and grading prostate cancer. For example, the infiltration degree of lung adenocarcinoma is judged based on the morphological analysis of multiple lesions. Therefore, self-attention-based methods more easily capture information related to the global context, resulting in correct judgments.

\begin{figure}[tbp]
\centering
\includegraphics[width=0.8\linewidth]{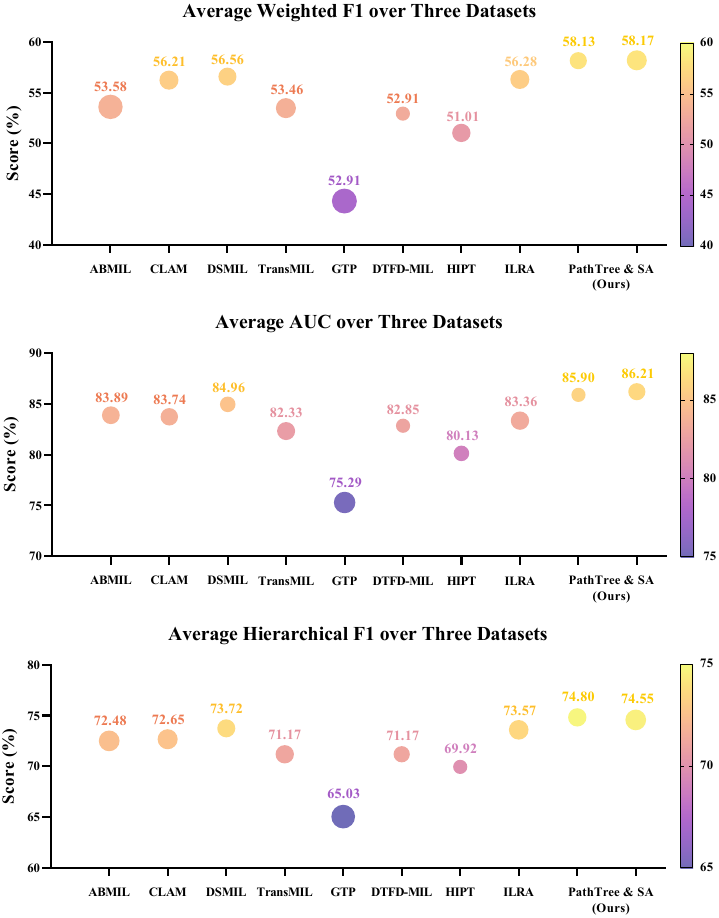}
\caption{Average results for the weighted F1, AUC, and H-F1 over three datasets.}
\label{overall_score}
\end{figure}

\begin{figure*}[htbp]
\centering
\includegraphics[width=0.77\linewidth]{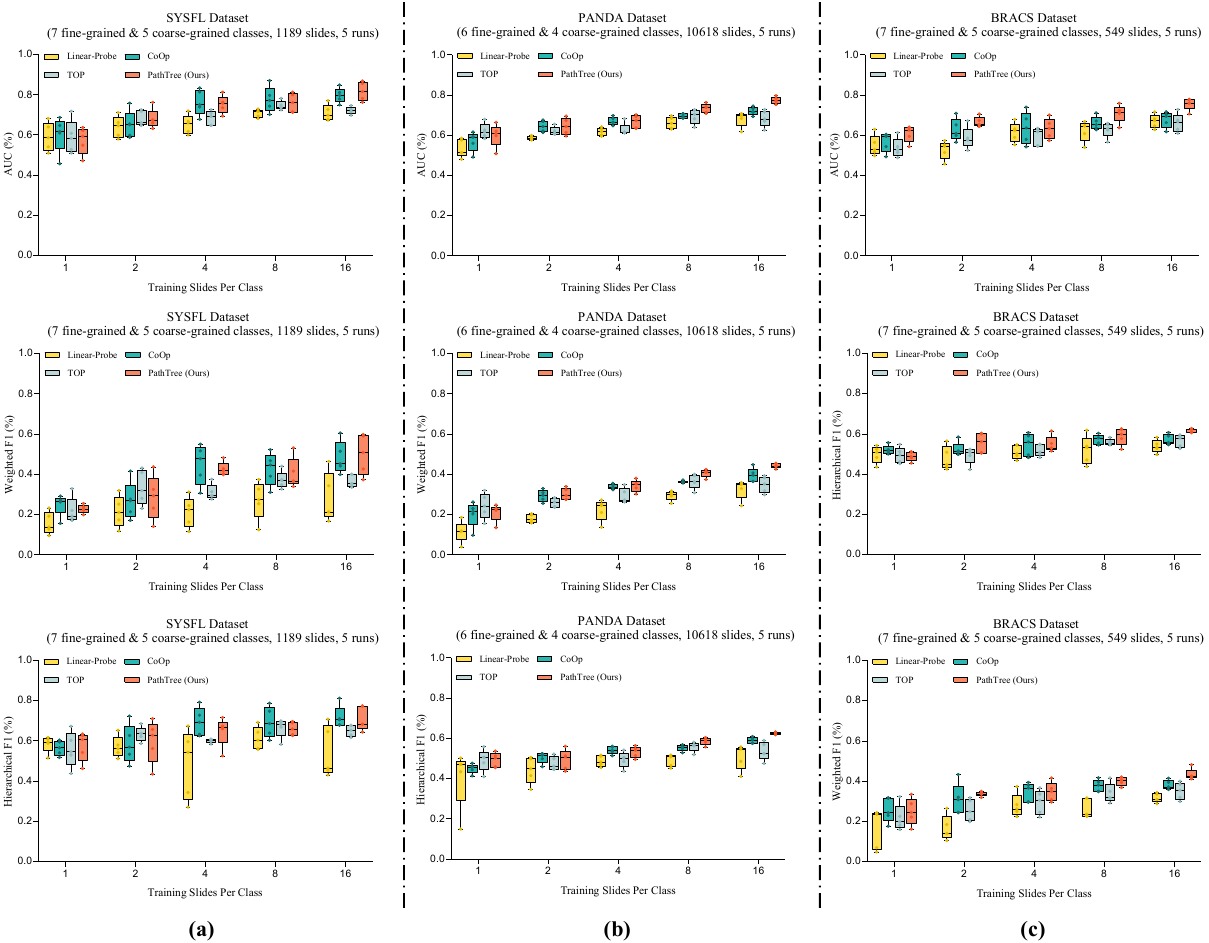}
\caption{AUC, Weighted F1, and H-F1 results [\%] on the fine-grained 7-category SYSFL, 6-category PANDA, and 7-category BRACS datasets in few-shot settings.}
\label{fewshot}
\end{figure*}

Table \ref{fine_hier} shows the hierarchical metric results for the three datasets. It should be noted that hierarchical metrics require considering both the coarse-to-fine prediction results and the ground truth labels for each slide. Our proposed PathTree methods achieve competitive levels in H-Precision and H-Recall, and outperform other methods by approximately 2\% in the more comprehensive H-F1 metric. This indicates that PathTree can distinguish slides more precisely than other state-of-the-art methods. Fig.\ref{overall_score} shows the average results for the weighted F1, AUC, and H-F1 metrics across the three datasets, which also confirms the superiority of PathTree-based methods at both planar and hierarchical levels.

\subsubsection{Coarse-grained Classification Results}

Hierarchical WSI analysis can be designed to address coarse-grained classification tasks based on pathological problems. SYSFL involves the study of frozen lung tissue sections. Lesions with good prognosis (AIS, AAH, and MIA) have similar properties and are low risk so that they can be clinically identified as one category; NOR and PNE are non-neoplastic and can be grouped into one category. Consequently, the coarse-grained SYSFL dataset is represented as a four-class classification task. PANDA focuses on grading prostate cancer, where an IUSP value greater than 0 indicates cancer. The distinction between the presence and absence of cancer is crucial for initial biopsy screening. Thus, the coarse-grained PANDA can be represented as a binary classification task. The BRACS dataset can be divided into three categories based on cancer risk: non-cancerous (NOR+BE+UDH), precancerous lesions (ADH+FEA), and cancerous (DCIS+IC). For each fold of fine-grained classification, models are saved based on the optimal weighted F1 score in the validation set and then directly inferred for each coarse-grained classification task. The related results are recorded in Table \ref{coarse_label}. It can be observed that although the accuracy for the BRACS dataset is slightly lower than the optimal result, the overall performance of the two PathTree methods remains superior.

\subsubsection{Comparison of Few-shot Learning} 

Contextual learning based on text prompts has been proven to have good few-shot capabilities (\citealt{radford2021learning,zhou2022conditional,zhou2022learning}). Fig.\ref{fewshot} shows the comparison results of PathTree with attention pooling-based Linear-Probe, CoOP, and TOP in 1, 2, 4, 8, and 16-shot settings. From the dataset perspective, PathTree exhibits the best few-shot capabilities on the BRACS and PANDA datasets, especially in BRACS, where it also has the smallest standard deviation. On the SYSFL dataset, CoOp also demonstrates strong performance. From the model capability perspective, Linear-Probe, which relies on fine-tuning a linear layer, performs poorly and has the largest standard deviation. However, CoOp and TOP, which incorporate text context tokens, improve performance.  From the shot settings perspective, although PathTree is not the best in the 1 and 2-shot, it shows effectiveness in the 4, 8, and 16-shot, which indicates that the fixed tree structure struggles to learn more contextual information with extremely few samples. However, as the number of samples slightly increases, the hierarchical structure can significantly learn the relationships within categories, thereby improving model performance.

\subsection{Ablation Experiments}

\subsubsection{Effectiveness of Different Text Prompts}

\begin{table}[tbp]
\caption{AUC results [\%] of PathTree with different text prompts on three fine-grained datasets.} 
\label{diff_text}
\centering
\resizebox{1.\columnwidth}{!}{
\renewcommand{\arraystretch}{1.3}{
\begin{tabular}{lccc}
\hline
Text prompt        & SYSFL          & PANDA          & BRACS        \\ \hline
'[CLASS]'  & $87.92_{1.63}$ & $87.25_{0.64}$ & $81.52_{2.30}$ \\
'this is [CLASS].'  & $87.69_{3.13}$ & $87.26_{0.50}$ & $80.66_{0.89}$ \\
'a photo of [CLASS].' & $87.69_{3.00}$ & $87.25_{0.13}$ & $81.63_{3.62}$ \\
'an image of [CLASS].'  & $87.22_{1.38}$ & $87.17_{0.36}$ & $80.29_{2.18}$ \\
'[CLASS] is present.'& $87.80_{2.48}$ & $87.28_{0.35}$ & $81.53_{2.44}$ \\
'a pathological image of [CLASS].' & $88.30_{1.94}$ & $87.24_{0.52}$ & $80.50_{2.47}$ \\
'a whole slide image of [CLASS].'  & $88.24_{2.59}$ & $87.33_{0.38}$ & $80.99_{2.54}$ \\ \hline
\textbf{Professional description (Ours)}  & $\mathbf{88.49_{2.21}}$ & $\mathbf{87.39_{0.47}}$ & $\mathbf{81.81_{2.43}}$\\  \hline
\end{tabular}
}}
\end{table}

Our proposed PathTree uses professional descriptions as prompt descriptions for learning and inference. To explore the impact of different text prompts on PathTree, we conduct experiments using different text prompts, including five conventional and two general pathology-related text prompts. Their comparative results are shown in Table \ref{diff_text}.We find that the performance of the two pathology-related texts is similar to that of other general text prompts but slightly better in the few-sample BRACS dataset, which suggests that specific semantic information can help the model extract more useful information, especially when the sample size is small. Moreover, our designed professional descriptions consistently outperformed all these methods. This indicates a strong potential correlation between the form of text prompts and model performance in PathTree, suggesting that problem-specific language may enhance the ability of text-guided PathTree in WSI analysis.

\subsubsection{Comparison of Different Graph-based Prompt Encoders}

\begin{figure}[tbp]
\centering
\includegraphics[width=1\linewidth]{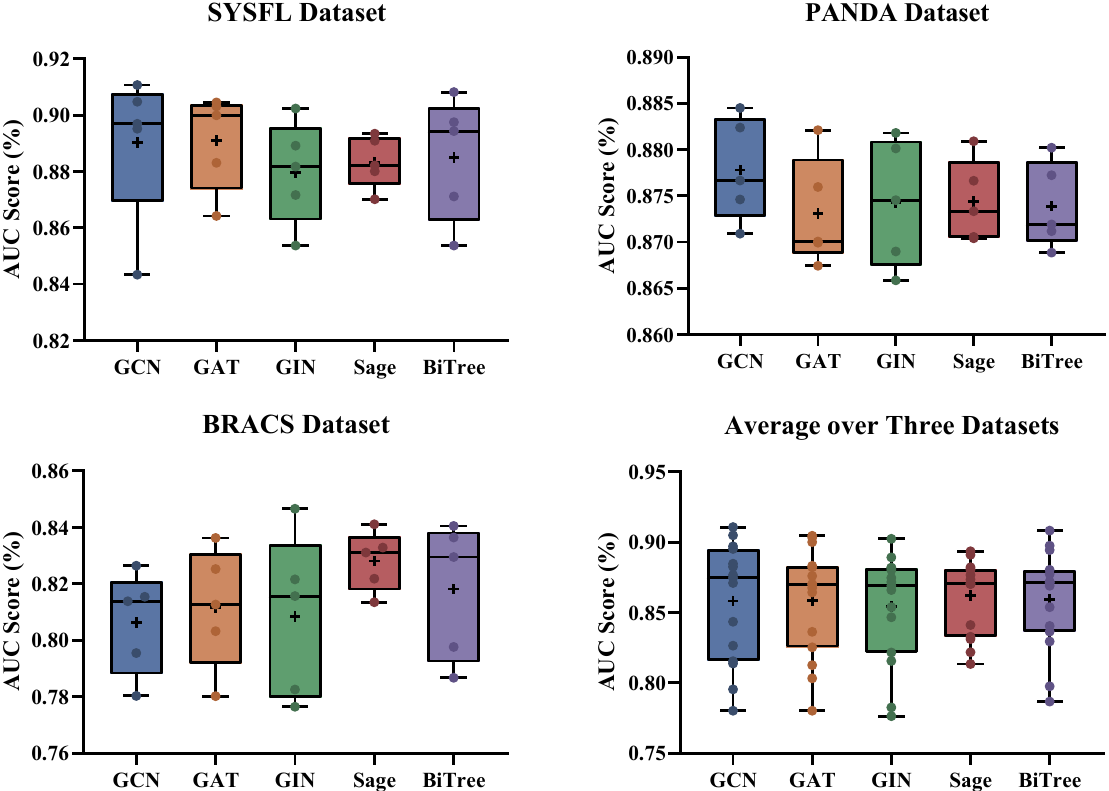}
\caption{AUC results [\%] with different graph-based text prompt encoders on three fine-grained datasets.}
\label{graph_tree}
\end{figure}

Text prompts pass through a tree structure to interact and convey more semantic information. In this paper, we model this tree structure as a graph representation and design a bidirectional directed graph structure (BiTree). We conducted ablation experiments comparing BiTree with several common graph architectures: graph convolutional network (GCN) (\citealt{kipf2017semisupervised}), graph attention network (GAT) (\citealt{veličković2018graph}), graph isomorphism network (GIN) (\citealt{xu2019powerfulgraphneuralnetworks}), and Sage (\citealt{sage}) to investigate their impact on model performance. The results are shown in Fig.\ref{graph_tree}. For the SYSFL dataset, GCN and BiTree perform better, with Sage showing less variance; for the PANDA dataset, GCN and GIN have slight advantages, but all methods show similar performance; for the BRACS dataset, Sage and BiTree are more competitive, with Sage showing less fluctuation. Overall, different graph-based text prompt encoders have little impact on the final performance of PathTree.

\subsubsection{Effect of Different Patch and Text Encoders}

\begin{table}[tbp]
\caption{AUC results [\%] of PathTree (text encoder: PLIP \citealt{huang2023visual}) with different patch encoders on three fine-grained datasets.}
\label{patch_encoder}
\centering
\resizebox{1.\columnwidth}{!}{
\renewcommand{\arraystretch}{1.3}{
\begin{tabular}{lccccccc}
\hline
Patch encoder & & & & & SYSFL & PANDA & BRACS \\ \hline
ImageNet (\citealt{deng2009imagenet}) & & & & & $82.90_{3.69}$    & $84.37_{0.87}$      & $74.53_{3.02}$      \\
KimiaNet (\citealt{riasatian2021fine}) & & & & & $84.03_{2.62}$    & $84.47_{0.31}$      & $74.60_{3.77}$  \\
CLIP (\citealt{radford2021learning})    & & & & & $85.97_{2.99}$    & $83.75_{0.89}$      & $74.47_{3.85}$        \\
PathDino (\citealt{Alfasly_2024_CVPR}) & & & & & $86.07_{2.99}$    & $89.05_{0.55}$      & $79.79_{2.90}$        \\
CONCH (\citealt{lu2024visual})    & & & & & $87.93_{1.96}$    & $\mathbf{89.94_{0.28}}$      & $\mathbf{85.60_{2.82}}$        \\
PLIP (\citealt{huang2023visual})     & & & & & $\mathbf{88.49_{2.21}}$    & $87.39_{0.47}$      & $81.81_{2.43}$        \\ \hline
\end{tabular}}}
\end{table}

\begin{table}[tbp]
\caption{AUC results [\%] of PathTree (patch encoder: PLIP \citealt{huang2023visual}) with different text encoders on three fine-grained datasets.}
\label{text_encoder}
\centering
\resizebox{1.\columnwidth}{!}{
\renewcommand{\arraystretch}{1.3}{
\begin{tabular}{lccccccc}
\hline
Text encoder & & & & & SYSFL & PANDA & BRACS \\ \hline
BERT (\citealt{kenton2019bert}) & & & & & $87.87_{1.97}$    & $87.30_{0.73}$      & $81.34_{2.80}$      \\
BioBERT (\citealt{lee2020biobert}) & & & & & $87.53_{3.41}$    & $87.26_{0.52}$      & $\mathbf{82.33_{1.84}}$  \\
CLIP (\citealt{radford2021learning})    & & & & & $87.99_{1.80}$    & $87.31_{0.74}$      & $81.28_{4.23}$        \\
CONCH (\citealt{lu2024visual})    & & & & & $87.89_{2.60}$    & $87.37_{0.46}$      & $81.86_{2.70}$       \\
PLIP (\citealt{huang2023visual})     & & & & & $\mathbf{88.49_{2.21}}$    & $\mathbf{87.39_{0.47}}$      & $81.81_{2.43}$        \\ \hline
\end{tabular}}}
\end{table}

Analyzing multimodal pathology vision-language data requires encoders for both images and text. We conduct further experiments to explore the impact of different patch and text encoders on the performance of PathTree. Table \ref{patch_encoder} presents the comparative results with various patch encoders while keeping the text encoder unchanged. Several conclusions can be drawn from these results. Firstly, the performance is poor when using patch encoders pre-trained on natural images, such as the ImageNet-supervised ViT-small or the CLIP pre-trained through image-text contrastive learning. This is likely due to the significant difference between natural and pathological image domains. 

Secondly, KimiaNet, obtained through fully supervised learning of pathological images, underperformed compared to PathDino, which is obtained through self-supervised learning based on the DINO (\citealt{caron2021emerging}). This suggests that current fully supervised learning paradigms may be less effective than self-supervised methods as general patch encoders for pathology-related downstream tasks, and the performance gap may be attributed to the smaller and less diverse training dataset KimiaNet uses.

Thirdly, patch encoders pre-trained with pathology image-text contrastive learning, such as PLIP and CONCH, perform better than those with purely visual supervision, such as KimiaNet and PathDino, which indicates that patch encoders pre-trained on multimodal data have greater potential as general image feature extractors for pathology tasks than purely visual encoders.

Table \ref{text_encoder} demonstrates the comparative results with various text encoders while keeping the image encoder unchanged. Encoders pre-trained on the biomedical text like BioBERT, CONCH, and PLIP perform slightly better than encoders pre-trained on natural language descriptions like BERT and CLIP. However, this advantage is not as large as the image encoders described above.

\subsubsection{Ablation Study of Tree-like Constraint Loss}

\begin{figure}[tbp]
\centering
\includegraphics[width=1\linewidth]{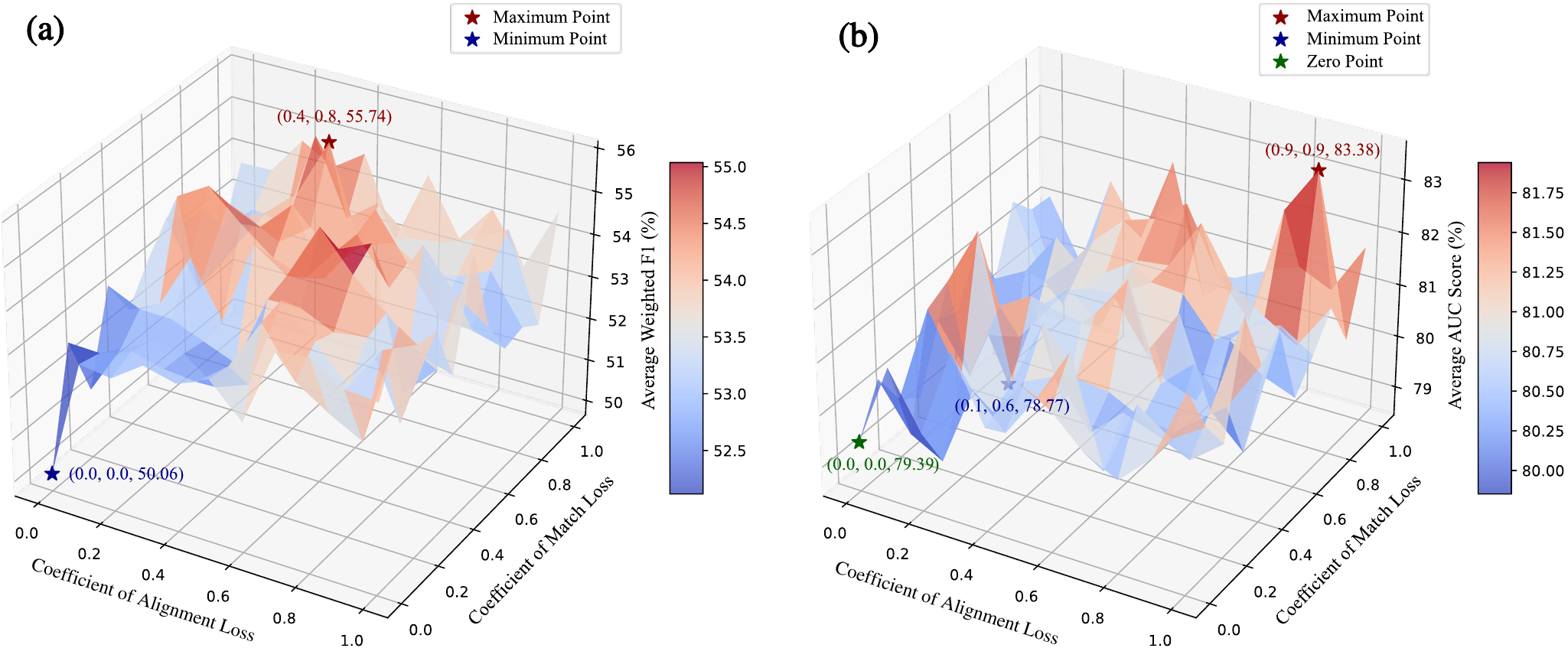}
\caption{The surface plots of PathTree under optimal average Weighted F1 and AUC score and different coefficient of Match Loss $\mu_{m}$, Alignment Loss $\mu_{p}$ in the BRACS dataset. The parameters $\mu_{m}, \mu_{p}$ range from $0.0$ to $1.0$ with step of $0.1$.}
\label{loss_3d}
\end{figure}

\begin{figure}[tbp]
\centering
\includegraphics[width=1\linewidth]{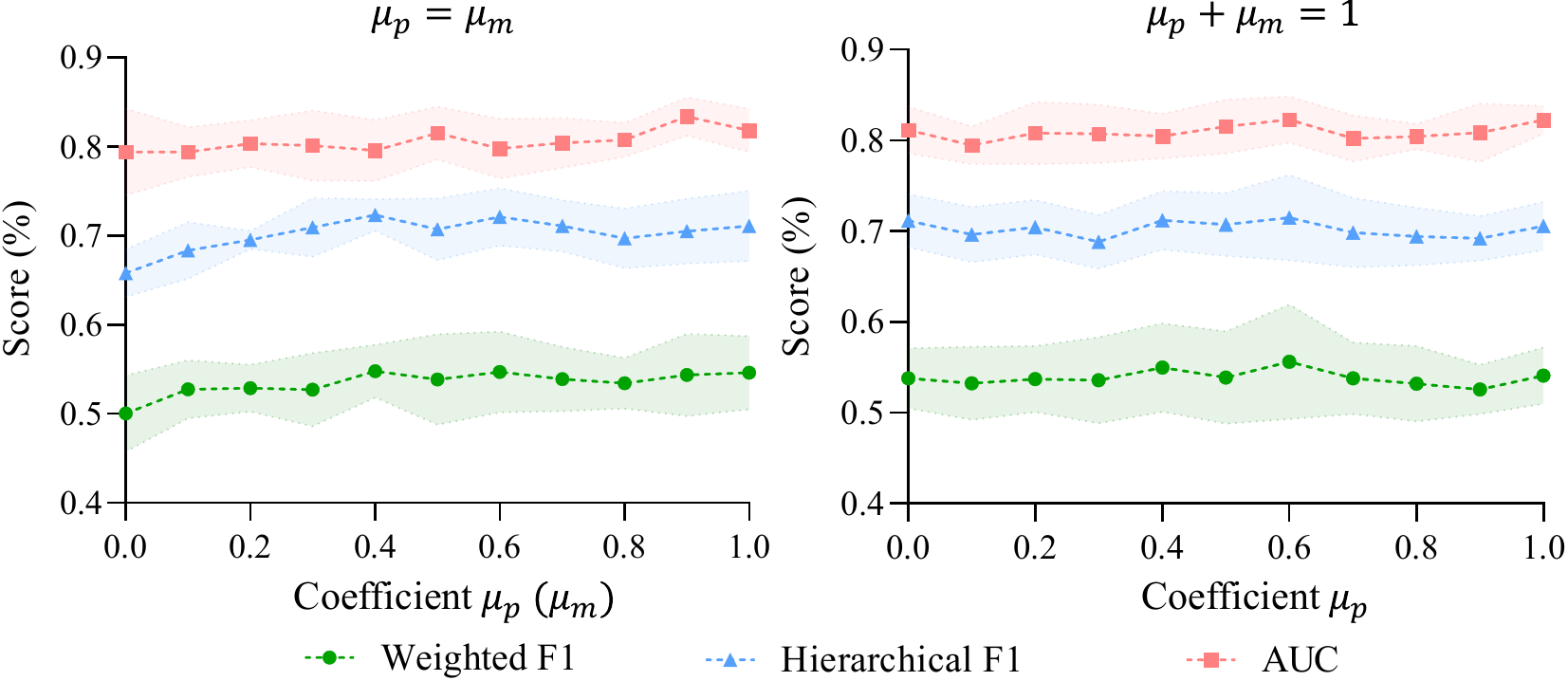}
\caption{Comparison of PathTree with different coefficients in the BRACS dataset. (a) Keep coefficients $\mu_p$ and $\mu_m$ equal. (b) Keep the sum of coefficients $\mu_p$ and $\mu_m$ equal to $1$.}
\label{loss_same_equal}
\end{figure}

Two joint constraint losses based on tree structures can guide the alignment of the slide with text embeddings, thereby affecting the analysis. Fig.\ref{loss_3d} shows the comparative results of PathTree on the BRACS dataset with different coefficients $\mu_{m}$ and $\mu_{p}$ for $\mathcal{L}_{match}$ and $\mathcal{L}_{path}$. For the Weighted F1 score, when neither $\mathcal{L}_{match}$ nor $\mathcal{L}_{path}$ is used, the performance of PathTree is the lowest, at only 50.06\%, which is 5.68\% lower than the highest result of 55.74\% obtained with these losses. For the AUC score, the worst result is at a non-zero point, but only 0.62\% lower than the zero point, and 4.61\% lower than the best result, demonstrating the effectiveness of the two multimodal alignment losses.

To further explore the potential relationship between constraint losses and model performance, we analyze two scenarios: when $\mu_m = \mu_p$ and when $\mu_m + \mu_p = 1$, which are shown in Fig.\ref{loss_same_equal}. Notably, when the two coefficients are always equal, there is a positive correlation between the coefficients and the three evaluation metrics. When the sum of the two coefficients is always 1, the results show a certain peak, indicating that the simultaneous use of both helps improve the performance. In summary, setting both coefficients relatively large and close to each other benefits PathTree. On the other hand, when these two constraint losses are not used, PathTree loses the strong correlation between slide and text embeddings, resulting in poorer performance.

\subsection{Visualization}

\begin{figure}[tbp]
\centering
\includegraphics[width=1\linewidth]{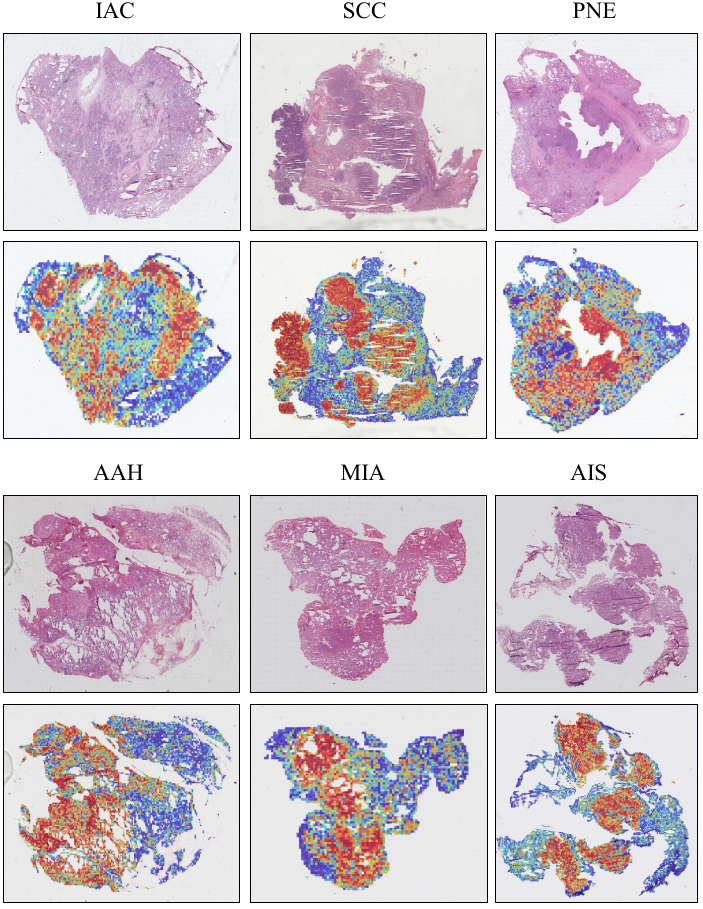}
\caption{Visualization heatmap results of PathTree in the SYSFL dataset.}
\label{heatmap}
\end{figure}

By visualizing the scores of corresponding categories in the multiple attention module on the patch regions, we can obtain WSI heatmaps of PathTree to demonstrate the interpretability. Fig.\ref{heatmap} shows six visualized samples of non-normal categories in the SYSFL dataset. For the IAC and SCC categories, PathTree can focus on extensive cancer regions. The visualization results for the PNE and AAH categories show that PathTree emphasizes lymphocytes and cell proliferation, which aligns well with the areas of interest in actual pathological diagnoses. For the MIA and AIS categories, the model highlights tumor cells growing along the walls and local lesions, which is critical for distinguishing these two in pathological morphology. 

\section{Conclusion}

In this paper, we transform conventional challenging pathological tasks from planar into tree-like analysis methods, introducing the concept of hierarchical WSI classification. Meanwhile, we propose a representation learning called PathTree for solving these hierarchical tasks. PathTree considers both text and WSI modalities, using professional pathology descriptions as coarse-grained and fine-grained prompts and introducing a tree-like graph to exchange semantic information between categories. To let the text guide WSI representation, PathTree aggregates multiple slide-level embeddings based on tree paths and uses slide-text similarity and additional metric losses to optimize the objectives of downstream tasks. In three challenging hierarchical classification tasks, we demonstrate that PathTree outperforms other state-of-the-art methods and can provide a more clinically relevant perspective for solving complex computational pathology problems.

\section*{Declaration of competing interest}
The authors declare that they have no known competing financial interests or personal relationships that could have appeared to influence the work reported in this paper.

\section*{CRediT authorship contribution statement}
\textbf{Jiawen Li: }Conceptualization, Investigation, Methodology, Data curation, Formal Analysis, Software, Validation, Visualization, Writing - original draft, Writing - review \& editing, Project administration. \textbf{Qiehe Sun: }Conceptualization, Methodology, Investigation, Software, Validation, Visualization, Writing - review \& editing. \textbf{Renao Yan: }Project administration, Software, Validation, Formal Analysis. \textbf{Yizhi Wang: }Data curation, Investigation, Validation. \textbf{Yuqiu Fu: }Data curation, Validation. \textbf{Yani Wei: }Resources, Data curation. \textbf{Tian Guan: }Project administration, Supervision. \textbf{Huijuan Shi: }Resources, Data curation, Project administration, Investigation, Supervision. \textbf{Yonghong He: }Project administration, Funding acquisition, Supervision. \textbf{Anjia Han: }Resources, Data curation, Writing – review \& editing, Supervision.

\section*{Acknowledgments}
This work is supported by the Shenzhen Engineering Research Centre (XMHT20230115004) and the Science and Technology Research Program of Shenzhen City (KCXFZ20201221173207022). We thank the support from the Jilin Fuyuan Guan Food Group Co., Ltd. In.

\bibliographystyle{model2-names.bst}\biboptions{authoryear}
\bibliography{refs}



\end{document}